\documentclass{article}

\usepackage{microtype}
\usepackage{graphicx}
\usepackage{subfigure}
\usepackage{booktabs} 
\usepackage{tabularx}

\usepackage{hyperref}

\usepackage[accepted]{icml2024}

\usepackage{amsmath}
\usepackage{amssymb}
\usepackage{mathtools}
\usepackage{amsthm}

\usepackage[capitalize,noabbrev]{cleveref}

\theoremstyle{plain}

\theoremstyle{definition}

\theoremstyle{remark}

\usepackage[textsize=tiny]{todonotes}

\begin{document}

\twocolumn[
\icmltitle{Leveraging Partial SMILES Validation Scheme for Enhanced Drug Design in Reinforcement Learning Frameworks}



\icmlsetsymbol{equal}{*}

\begin{icmlauthorlist}
\icmlauthor{Xinyu Wang}{uconn}
\icmlauthor{Jinbo Bi}{uconn}
\icmlauthor{Minghu Song}{hefei}
\end{icmlauthorlist}

\icmlaffiliation{uconn}{Department of Computer Science \& Engineering, University of Connecticut, Connecticut, USA}
\icmlaffiliation{hefei}{Hefei Comprehensive National Science Center, Hefei, China}

\icmlcorrespondingauthor{Xinyu Wang}{xinyu.wang@uconn.edu}

\icmlkeywords{Machine Learning, ICML,Drug Discovery}

\vskip 0.3in
]



\printAffiliationsAndNotice{}  
\begin{abstract}
SMILES-based molecule generation has emerged as a powerful approach in drug discovery. Deep reinforcement learning (RL) using large language model (LLM) has been incorporated into the molecule generation process to achieve high matching score in term of likelihood of desired molecule candidates. However, a critical challenge in this approach is catastrophic forgetting during the RL phase, where knowledge such as molecule validity, which often exceeds 99\% during pretraining, significantly deteriorates. Current RL algorithms applied in drug discovery, such as REINVENT, use prior models as anchors to retian pretraining knowledge, but these methods lack robust exploration mechanisms. To address these issues, we propose Partial SMILES Validation-PPO (PSV-PPO), a novel RL algorithm that incorporates real-time partial SMILES validation to prevent catastrophic forgetting while encouraging exploration. Unlike traditional RL approaches that validate molecule structures only after generating entire sequences, PSV-PPO performs stepwise validation at each auto-regressive step, evaluating not only the selected token candidate but also all potential branches stemming from the prior partial sequence. This enables early detection of invalid partial SMILES across all potential paths. As a result, PSV-PPO maintains high validity rates even during aggressive exploration of the vast chemical space. Our experiments on the PMO and GuacaMol benchmark datasets demonstrate that PSV-PPO significantly reduces the number of invalid generated structures while maintaining competitive exploration and optimization performance. While our work primarily focuses on maintaining validity, the framework of PSV-PPO can be extended in future research to incorporate additional forms of valuable domain knowledge, further enhancing reinforcement learning applications in drug discovery.
\end{abstract}
%

\section{Introduction}
The Simplified Molecular Input Line Entry System (SMILES) \cite{weininger1988smiles} has been widely adopted as one of molecular representations for various molecular generation and prediction tasks \cite{segler2018generating,bjerrum2017molecular,irwin2022chemformer,bagal2021molgpt}.
Large language model (LLM) have demonstrated strong capabilities in SMILES-based molecule generation, and reinforcement learning (RL) is often employed to fine-tune these models to optimize desired molecular properties \cite{mokaya2023testing,korshunova2022generative}.
\begin{figure*}[htbp]
\centering
\includegraphics[width=1.\textwidth]{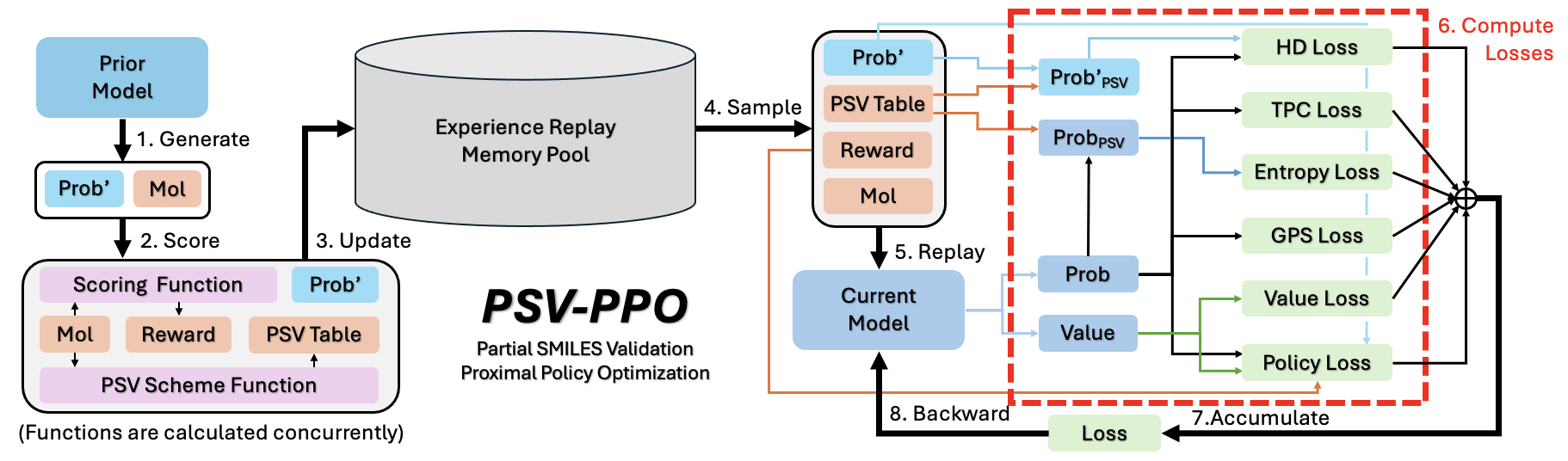}
\caption{The training flow of the PSV-PPO algorithm. The process begins with the prior model generating molecular structures and their associated probabilities (1). Both the rewards and the Partial SMILES Validation (PSV) Table are computed concurrently, which minimizes the computational overhead (2). The experience replay memory pool is updated with the scored molecules (3). Samples are then drawn from the experience replay memory pool (4) and replayed through the current model (5). The PSV-PPO algorithm calculates multiple loss functions, including HD Loss, TPC Loss, Entropy Loss, GPS Loss, Value Loss, and Policy Loss (6), which are then accumulated (7) and used to perform a backward pass to update the model parameters (8). The PSV scheme ensures that the model learns to generate valid molecular structures by providing real-time feedback on token validity during the generation process.}
\label{fig:overview}
\end{figure*}

Despite their strong performance during pretraining, language models often experience catastrophic forgetting when fine-tuned with RL\cite{kirkpatrick2017overcoming}. Reinforce Learning with exploration, while widely used across various applications, frequently overrides the model’s prior knowledge\cite{kemker2018measuring}, leading to instability, increased generation of invalid molecules, and inefficient learning. This issue is particularly concerning in molecular generation, where essential properties such as validity, which can exceed 99\% during pretraining\cite{arus2020smiles}, deteriorate significantly once RL optimization begins.

To address pretraining knowledge degradation during RL, several methods have been proposed. Reinvent \cite{blaschke2020reinvent} uses a pre-trained prior model to maintain learned knowledge while guiding RL-based optimization. While this approach helps retain molecular validity, it limits the ability of the model to explore novel chemical structures, restricting diversity. Other methods, such as SELFIES \cite{krenn2022selfies} and grammar-VAE \cite{kusner2017grammar}, enforce validity at the representation level by constraining molecular embeddings. Although these techniques effectively eliminate invalid structures at the encoding stage, studies have shown that SMILES-based methods often outperform them in terms of molecular property optimization and diversity \cite{ghugare2023searching}. The rigid constraints imposed by these representations can hinder efficient chemical space exploration, reducing their effectiveness in real-world drug discovery applications. 

Among RL algorithms, proximal policy optimization (PPO) has been widely adopted due to its strong performance in various domains, including natural language processing and control tasks\cite{wu2023brief}. However, its application to SMILES-based molecular generation presents significant challenges. The entropy-driven exploration mechanism in PPO encourages diverse sampling but can lead to instability\cite{hao2023exploration}. Large policy updates introduce excessive variation between consecutive iterations, causing gradient explosion, which destabilizes training. On the other hand, reducing entropy to maintain stability leads to mode collapse, where the model consistently selects the same high-probability tokens\cite{shi2018toward}. This issue is particularly problematic in SMILES-based generation, where token dependencies are strict, and minor errors can render an entire molecule invalid. A collapsed policy may excessively favor frequently occurring tokens, disrupting structural integrity and reducing the validity of generated molecules. The challenge of balancing exploration and exploitation in this setting makes PPO-based molecular generation highly unstable. 

Another fundamental issue in RL-based molecule generation is sparse and delayed rewards. Molecular validity and property scores are typically assessed only after a full SMILES string is generated, preventing the model from receiving real-time feedback on structural correctness. Without intermediate guidance, PPO fine-tuning may lead the model toward structurally incorrect molecules, resulting in inefficient learning. Some methods embed validity constraints directly into the generation process, such as SELFIES and grammar-enforced models, but these approaches have been found to be less effective than elfSMILES-based techniques in optimizing molecular properties \cite{ghugare2023searching}. Without an intermediate validation mechanism, PPO fine-tuning degrades molecular validity over time, leading to inefficient exploration and reduced performance in molecular optimization tasks. 

To overcome these challenges, we propose \textbf{P}artial \textbf{S}MILES \textbf{V}alidation - \textbf{P}roximal \textbf{P}olicy \textbf{O}ptimization (\textbf{PSV-PPO}), a reinforcement learning framework that integrates real-time molecular validity checks into PPO. Unlike conventional RL approaches that evaluate validity only after generating a full molecule, PSV-PPO introduces a partial SMILES validation mechanism that dynamically penalizes invalid token selections at each autoregressive step. This real-time intervention prevents errors from propagating, stabilizing training while improving overall molecular validity. 

The core of PSV-PPO is the PSV truth table, which systematically evaluates candidate tokens before they are appended to a growing SMILES string. If a token is determined to lead to an invalid structure, the policy network is immediately penalized. This approach ensures:
\begin{itemize}
    \item \textbf{Immediate feedback} during generation, reducing wasted exploration.
    \item \textbf{Controlled exploration}, allowing for diverse yet syntactically valid molecule generation.
    \item \textbf{Stable training} by mitigating mode collapse and gradient explosion.
\end{itemize}
Our main contributions are as follows:
\begin{enumerate}
    \item \textbf{Partial SMILES Validation Mechanism:} We introduce the PSV truth table, a principled approach to verify token validity before appending it to a growing SMILES sequence. While PSV does not guarantee a completely valid final molecule, it ensures that invalid token sequences are detected and penalized during generation.
    
    \item \textbf{PSV-PPO Algorithm:} We seamlessly integrate PSV into PPO, enabling effective molecule generation while maintaining exploration and stability. This approach significantly improves molecule validity while preventing degenerative policies.
    
    \item \textbf{Adaptive Loss Regularization:} We introduce new loss terms that dynamically penalize tokens with excessively high probabilities, preventing mode collapse. The PSV table determines when to apply this regularization, ensuring effective exploration while maintaining token validity.
\end{enumerate}

\section{Related Works}
\subsection{Machine Learning for Drug Discovery}
The integration of machine learning into drug discovery has significantly transformed the landscape of molecular design, enabling the generation of novel compounds through various molecular representations, including 1D, 2D, and 3D formats \cite{wu2018moleculenet,Ramsundar-et-al-2019,lo2018machine}. These approaches utilize advanced deep learning architectures to model and predict chemical properties, facilitate molecule generation, and optimize lead compounds.

\paragraph{SMILES Representation.} SMILES is a text-based notation that encodes molecular structures as sequences of characters, where each character or set of characters corresponds to specific atoms, bonds, or topological features. This linear representation simplifies the complex task of molecule generation into a sequence prediction problem, similar to those in natural language processing (NLP). Leveraging this analogy, models such as recurrent neural networks (RNNs) and transformers, originally developed for NLP tasks, have been adapted to generate and analyze SMILES strings \cite{gomez2018automatic,jin2018junction}.


\paragraph{2D and 3D Molecule Generation.} While SMILES strings provide a straightforward and compact representation, 2D and 3D molecular representations offer more detailed insights into a molecule's spatial and topological properties, which are crucial for understanding molecular interactions, predicting chemical properties, and designing drugs with specific biological activities. Machine learning models, particularly those based on graph neural networks (GNNs) and spatial convolutional networks, have been increasingly applied to generate and optimize these 2D and 3D structures. These models capture the complex geometry and electronic distribution of molecules, providing a more nuanced approach to drug discovery \cite{duvenaud2015convolutional,gilmer2017neural}.

\paragraph{Atom-wise Tokenization.} Atom-wise tokenization offers an alternative to traditional SMILES-based representations by breaking down molecules into individual atoms and their associated bonds. This finer granularity enables models to better capture subtle variations in molecular structures that might be overlooked in more coarse-grained representations. By focusing on the atom-level details, this approach improves the accuracy and diversity of generated molecules, enhancing the model's ability to predict and optimize molecular properties \cite{kearnes2016molecular,jin2020hierarchical}.

\paragraph{Partial SMILES.} The \texttt{partialsmiles} package \cite{partialsmiles2024} provide real-time syntax validation, kekulization, and valence checks, allowing for on-the-fly detection of errors in SMILES strings as they are generated. The Partial SMILES performs three key checks:
\begin{enumerate}
    \item \textbf{Syntax Compliance:} Guarantees that the SMILES string follows the established syntactical conventions.
    \item \textbf{Aromaticity Handling:} Checks if aromatic systems can be appropriately kekulized by making implicit bonds explicit where required.
    \item \textbf{Valence Validation:} Ensures that the valence of each atom falls within the acceptable chemical range.
\end{enumerate}
\subsection{Revisit of PPO}
PPO is a widely adopted RL algorithm that enhances the stability of policy gradient methods through the use of a surrogate objective function \cite{schulman2017proximal}. PPO is commonly implemented using frameworks such as OpenAI Baselines \cite{baselines}, which provide standard implementations of various reinforcement learning algorithms. This carefully balanced objective enables PPO to improve policies while maintaining stability, making it a robust and reliable choice for a wide range of reinforcement learning tasks, including complex robotic control \cite{andrychowicz2020learning}. PPO builds on the principles of Trust Region Policy Optimization (TRPO) \cite{schulman2015trust}, carefully controlling the magnitude of policy updates, ensuring that the new policy remains within a constrained region close to the old policy, thereby improving learning stability and efficiency \cite{henderson2018deep}.

\begin{figure*}[htbp]
\centering
\includegraphics[width=1.\textwidth]{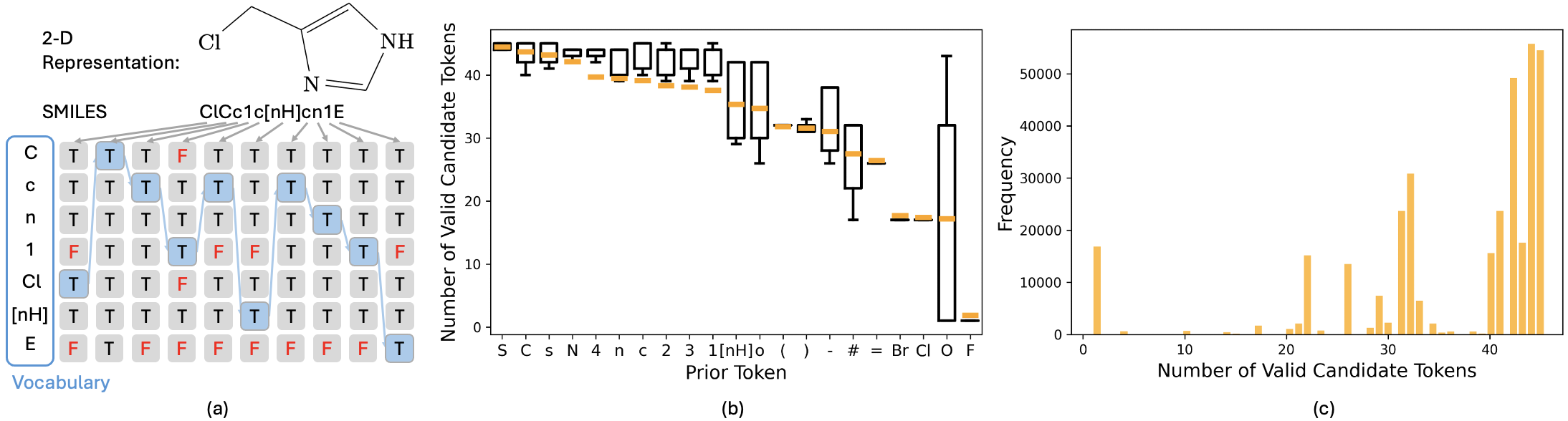}
\caption{Visualization of the Partial SMILES Validation (PSV) framework used in our study. \textbf{(a)} PSV Truth Table illustrating the validity assessment of each candidate token in a partially generated SMILES string. \textbf{(b)} Box plot showing the distribution of the number of valid candidate tokens based on the preceding token in the SMILES string. \textbf{(c)} Overall distribution of valid candidate tokens across all states derived from 10,000 sampled molecules.}
\label{fig:partial}
\end{figure*}

At the core of PPO is the clipped surrogate objective, designed to maximize expected advantage while constraining policy changes to prevent overly large updates. Given a policy \(\pi_{\theta}\) parameterized by \(\theta\), the probability ratio \(r_t(\theta)\) between the new and old policies is defined as:
\begin{equation}
r_t(\theta) = \frac{\pi_{\theta}(a_t | s_t)}{\pi_{\theta_{\text{old}}}(a_t | s_t)}    
\end{equation}
The objective function for PPO is then expressed as:
\begin{equation}
\begin{split}
L^{\text{CLIP}}(\theta) &= \mathbb{E}_t \Big[ \min \big( r_t(\theta) A(s_t, a_t), \\
&\quad \text{clip}(r_t(\theta), 1 - \epsilon, 1 + \epsilon) A(s_t, a_t) \big) \Big]    
\end{split}
\end{equation}
where \(A(s_t, a_t)\) represents the advantage estimate, and \(\epsilon\) is a hyperparameter controlling the clip range. This clipped objective mitigates the risk of excessive policy divergence, striking a balance between exploration and exploitation.

In addition to the policy loss, PPO optimizes a value function \(V_{\phi}(s_t)\) by minimizing the squared error between predicted values and actual returns, denoted as:
\begin{equation}
L^{\text{VALUE}}(\phi) = \mathbb{E}_t \left[ \left( V_{\phi}(s_t) - R_t \right)^2 \right]    
\end{equation}
where \(R_t\) is the cumulative return at time step \(t\).

To encourage sufficient exploration, PPO incorporates entropy regularization, penalizing overly deterministic policies through an entropy loss:
\begin{equation}
L^{\text{ENTROPY}}(\theta) = -\beta \mathbb{E}_t \left[ \sum_a \pi_{\theta}(a | s_t) \log \pi_{\theta}(a | s_t) \right]  
\end{equation}
where \(\beta\) is the entropy coefficient that controls the regularization strength.

Moreover, a Kullback-Leibler (KL) divergence \cite{kullback1951information} term can be optionally included to further restrict the policy update, ensuring the updated policy does not deviate significantly from the previous one:
\begin{equation}
L^{\text{KL}}(\theta) = \mathbb{E}_t \left[ \text{KL} \left[ \pi_{\theta_{\text{old}}}(\cdot | s_t) \parallel \pi_{\theta}(\cdot | s_t) \right] \right]    
\end{equation}
The overall objective function in PPO is a weighted combination of these components:
\begin{equation}
L(\theta, \phi) = L^{\text{CLIP}}(\theta) + c_1 L^{\text{VALUE}}(\phi) - c_2 L^{\text{ENTROPY}}(\theta) + c_3 L^{\text{KL}}(\theta)    
\end{equation}
where \(c_1\), \(c_2\), and \(c_3\) are coefficients that balance the contributions of the value loss, entropy loss, and KL divergence, respectively. This carefully balanced objective enables PPO to effectively improve policies while maintaining stability, making it a robust and reliable choice for a wide range of reinforcement learning tasks, including complex robotic control \cite{andrychowicz2020learning}. Practical implementations of PPO, such as those provided by Kostrikov \cite{kostrikov2018pytorch}, are widely used in the reinforcement learning community for research and application.

\section{Partial SMILES Validation PPO (PSV-PPO)}
The overall architecture of PSV-PPO is depicted in Figure \ref{fig:overview}, which illustrates how PSV-PPO builds upon the principles of Proximal Policy Optimization (PPO) while incorporating the Partial SMILES Validation (PSV) table into the training loop. We first explain how we build a PSV table given a generated SMILES string, then 

The core innovation of PSV-PPO lies in its ability to ensure chemically valid exploration of the molecular space. This is achieved through the integration of the PSV table, which plays a crucial role in guiding the model away from invalid token sequences during the training process. Additionally, PSV-PPO introduces several novel loss functions: PSV-Driven Entropy Loss, PSV-Driven Hellinger Distance Loss, Global Probability Stabilization Loss (GPS Loss), and Token Probability Control Loss (TPC Loss). These losses collectively prevent the model from overcommitting to specific molecular structures or tokens, thereby promoting diversity in the generated molecules.

\subsection{PSV Truth Table}
The PSV table is the foundational component of the PSV-PPO framework. The goal is to identify the validity of each candidate token at every step of the tokenized SMILES sequence generation. Figure \ref{fig:partial} (a) represents an PSV table for the molecule "ClCc1c[nH]cn1". The tokenized sequence of this candidate tokens are shown as then row of the matrix table. Each column in the table is the binary vector for all candidate tokens from the vocabulary and its value indicates whether the current token will immediately cause the partially generated SMILES become invalid. The pseudo-code for generating the PSV table is provided in the Appendix.

\subsection{Objectives of PSV-PPO}
The PSV-PPO frameowork, illustrated in Figure\ref{fig:overview}, extends the standard PPO algorithm and incorporates experience replay domain-specific enhancements. At the beginning of one PSV-PPO learning epoch, molecular samples are initially generated by the prior policy, and these samples, along with their associated rewards and PSV tables, are stored in an experience replay memory pool. To optimize computational efficiency, we the probability distribution of all actions from the prior model is also stored in the memory pool. This stored probability distribution contains not only the probability of action tokens, but also all other candidates tokens at each state. 

The current policy then replays molecules sampled from the memory pool, generating a probability distribution for the actions with the same way as the prior policy. Concurrently, the critic network predicts values that contribute to the calculation of the value loss. While the clip loss and value loss are retain their original forms, the entropy loss and KL divergence loss are modified to integrate domain specific knowledge from the PSV table into the learning process. Additionally, two new loss functions are introduced to mitigate the risk of mode collapse, particularly when the model struggles to discover molecules with higher scores than those already present in the memory pool. The final loss is a composite of the six sub-losses mentioned above, and it is used to perform backpropagation, thereby completing one epoch of the PSV-PPO learning process.
\subsubsection{PSV-Driven Entropy Loss}
The PSV-Driven Entropy Loss leverages the insights provided by the Partial SMILES Validation Grid (PSV table) to refine the traditional entropy regularization term. Traditional entropy regularization encourages exploration by maximizing the entropy of the action distribution, but it does so indiscriminately, potentially guiding the model toward chemically invalid states. The PSV-Driven Entropy Loss restricts exploration to tokens that lead to valid SMILES substrings while preventing the model from favoring actions with larger valid token sets.
\begin{equation}
L^{\text{ENTROPY}}_{PSV}(\theta) = -\beta \mathbb{E}_t \left[ \sum_{a \in D_{PSV}} \frac{\pi_{\theta}(a | s_t) \log \pi_{\theta}(a | s_t)}{\log(\text{len}(D_{PSV}))} \right]
\end{equation}

Here, the PSV-valid vocabulary $D_{PSV}$ consists of tokens validated by the PSV table, given the partially generated SMILES string at state $s_t$. The normalization factor, $\log(\text{len}(D_{PSV}))$, ensures that the entropy loss remains independent of the size of the valid token set, preventing the model from being biased towards actions with larger valid token sets and maintaining balanced exploration. Our experiments show that without this normalization, the model tends to converge prematurely to sub-optimal states, particularly favoring tokens associated with larger valid action spaces, such as the non-aromatic carbon token "C". By normalizing the entropy loss, the exploration process is guided by chemical validity rather than token set size, fostering more robust and diverse molecule generation.

\subsubsection{PSV-Driven Hellinger Distance Loss}
While the PSV-Driven Entropy Loss encourages the selection of valid tokens, it does not explicitly penalize the model for considering invalid tokens. To address this, we adapt the Hellinger Distance (HD) loss \cite{hellinger1909neue,arjovsky2017wasserstein}. Traditionally, KL divergence is used to measure the difference between the current policy \(\pi_{\theta}\) and the prior policy \(\pi_{\theta_{\text{old}}}\), regularizing the learning process. However, KL divergence is incompatible with \(\pi_{\theta_{\text{old\_PSV}}}\) because it contains zero probabilities introduced by the PSV table.

In our approach, we replace KL divergence with the Hellinger Distance, which is better suited for cases where the prior policy’s output is filtered through the PSV table, excluding invalid tokens. The PSV-Driven Hellinger Distance Loss is defined as:
\begin{equation}
L^{\text{HD}}_{PSV}(\theta) = \mathbb{E}_t \left[ \text{HD} \left[ \pi_{\theta_{\text{old\_PSV}}}(\cdot | s_t) \parallel \pi_{\theta}(\cdot | s_t) \right] \right]
\end{equation}
Here, \(\pi_{\theta_{\text{old\_PSV}}}\) represents the prior policy’s distribution filtered by the PSV table, ensuring that only valid actions are considered. This targeted penalty encourages the model not only to favor valid tokens but also to learn from and correct potential errors, promoting the generation of valid SMILES strings while maintaining balanced exploration of the chemical space.
\begin{table*}[h]
\centering
\caption{Scores of PSV-PPO and other baseline models on the GuacaMol benchmark.}
\label{tab:benchmark_scores}
\footnotesize  
\begin{tabularx}{\textwidth}{lcccccccc}
\hline
\hline
\textbf{Tasks}                        & \textbf{SMILES GA} & \textbf{SMILES LSTM} & \textbf{Graph GA} & \textbf{Reinvent} & \textbf{GEGL} & \textbf{MolRL-MGPT} & \textbf{PSV-PPO} \\ \hline
C\textsubscript{11}H\textsubscript{24}                 & 0.829              & 0.993                & 0.971             & 0.999             &\textbf{1.000}        & \textbf{1.000}               & \textbf{1.000}                \\ 
C\textsubscript{9}H\textsubscript{10}N\textsubscript{2}O\textsubscript{2}P\textsubscript{2}Cl  & 0.889              & 0.879                & 0.982             & 0.877             &\textbf{1.000}        & 0.939               & \textbf{1.000}                \\ \hline
Median molecules 1                & 0.334              & 0.438                & 0.406             & 0.434             & 0.455         & 0.449               & \textbf{0.459}                \\ 
Median molecules 2               & 0.380              & 0.422                & 0.432             & 0.395             & \textbf{0.437}         & 0.422               & 0.392                \\ \hline
Osimertinib MPO                  & 0.886              & 0.907                & 0.953             & 0.889             &\textbf{1.000}        & 0.977               & 0.951                \\ 
Fexofenadine MPO                 & 0.931              & 0.959                & 0.998             &\textbf{1.000}            &\textbf{1.000}        &\textbf{1.000}              & \textbf{1.000}                \\ 
Ranolazine MPO                   & 0.881              & 0.855                & 0.920             & 0.895             & 0.933         & \textbf{0.939}               & 0.936               \\ 
Perindopril MPO                  & 0.661              & 0.808                & 0.792             & 0.764             & 0.833         & 0.810               & \textbf{0.849}                  \\ 
Amlodipine MPO                   & 0.722              & 0.894                & 0.894             & 0.888             & 0.905         & 0.906               & \textbf{0.908}                \\ 
Sitagliptin MPO                  & 0.689              & 0.545                & \textbf{0.891}             & 0.539             & 0.749         & 0.823               & 0.804                \\ 
Zaleplon MPO                     & 0.413              & 0.669                & 0.754             & 0.590             & 0.763         & \textbf{0.790}               & 0.764                \\ \hline
Valsartan SMARTS                 & 0.552              & 0.978                & 0.990             & 0.095             &\textbf{1.000}        & 0.997               & 0.999                \\ \hline
\hline
\end{tabularx}
\end{table*}
\subsubsection{Final Loss Function}
The final loss function integrates these key components into the overall optimization process:
\begin{equation}
\begin{split}
    Loss & = L^{\text{CLIP}}(\theta) + L^{\text{Value}}(\theta) + L^{\text{ENTROPY}}_{PSV}(\theta)\\
    & + L^{\text{HD}}_{PSV}(\theta) + L^{\text{TPC}}_{PSV}(\theta) + L^{\text{GPS}}_{PSV}(\theta)  
\end{split}
\end{equation}
where $L^{\text{TPC}}_{PSV}(\theta)$ and $L^{\text{GPS}}_{PSV}(\theta)$ are additional regularization terms designed to prevent mode collapse. Detailed explanations can be found in the Appendix.
\begin{figure*}[htbp]
\centering
\includegraphics[width=0.95\textwidth]{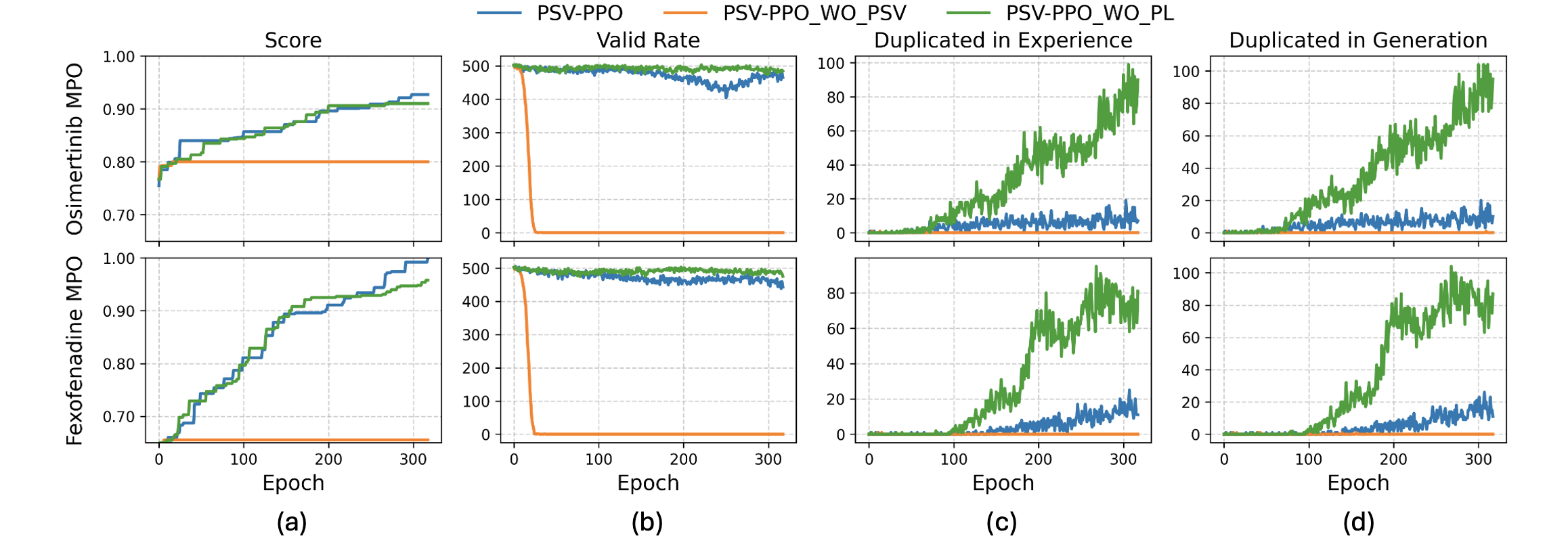}
\caption{Ablation study results for the Fexofenadine MPO and Osimertinib MPO tasks, illustrating the impact of various components within the PSV-PPO framework. The plots compare PSV-PPO with two ablation models: PSV-PPO\_WO\_PSV, which uses the standard PPO entropy and KL divergence losses without integrating the PSV table, and PSV-PPO\_WO\_PL, which excludes the GPS and TPC losses. The score plots demonstrate the optimization effectiveness of each model across epochs. The valid rate plots reveal how the absence of the PSV table significantly reduces the validity of generated molecules. The duplication plots in both the experience replay memory and during generation highlight the role of GPS and TPC losses in maintaining diversity and preventing mode collapse in the generated molecules.}
\label{fig:ablation}
\end{figure*}
\begin{figure}[htbp]
\centering
\includegraphics[width=0.40\textwidth]{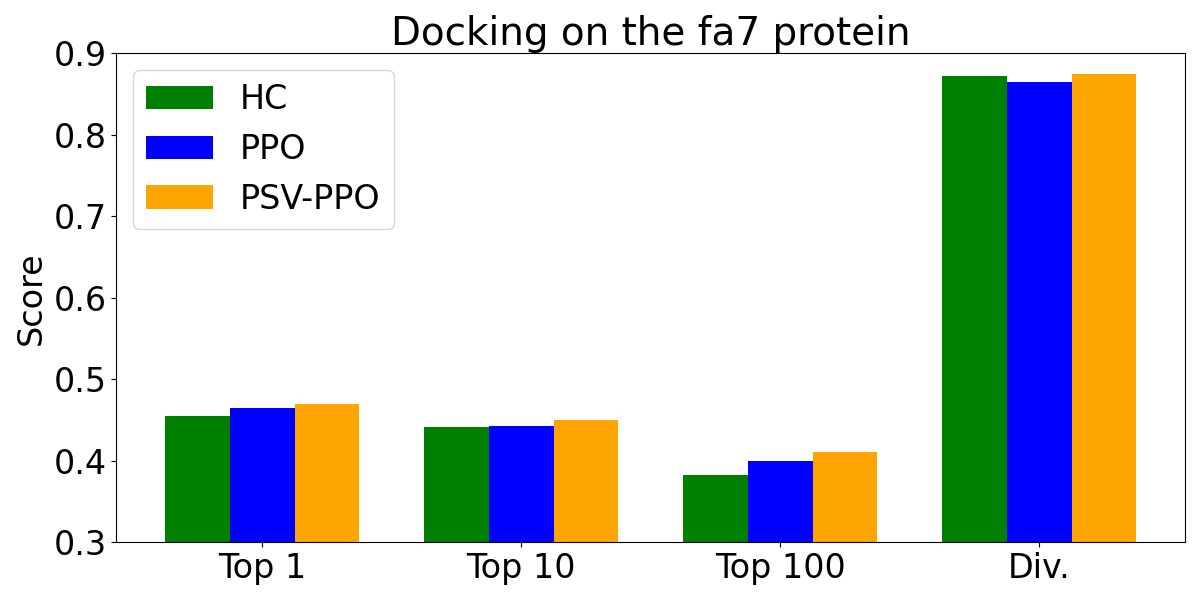}
\caption{Docking Performance: Comparison of molecular docking performance on the fa7 target across different reinforcement learning methods. The plot presents scores for Top 1, Top 10, Top 100, and Diversity (Div.), evaluating HC (green), PPO (blue), and PSV-PPO (orange). PSV-PPO achieves competitive performance, demonstrating its effectiveness in generating diverse and high-quality molecules.}
\label{fig:docking}
\end{figure}
\section{Experimental Results}
\subsection{PMO Benchmark}
The PMO benchamrk \cite{gao2022sample} evaluates molecular generation models using the area under the curve (AUC) of the top-K average property value versus the number of oracle calls. Following previous studies, we report results for $K=10$. To ensure fair evaluation, we limit the number of oracle calls to $10,000$ and allow early stopping for HC and PPO-based methods to minimize the total number of oracle calls required to achieve the best scores. We implemented PSV-PPO on the pretrained LSTM model provided by the benchmark, named LSTM PSV-PPO. The pretrained LSTM model consists of 6 layers with 512 hidden dimension, and a dropout rate of 0.2. The cross entropy loss achieved by the pretrained model is 0.407.

We compare our method with Reinvent, LSTM Hill Climbing (LSTM HC), LSTM PPO and SMILES PSV-PPO. LSTM HC is an iterative learning method that fine-tunes a generative LSTM model by incorporating high-scoring molecules into the training set at each iteration. As a variant of the cross-entropy method and REINFORCE with specific reward shaping, LSTM HC adaptively refines the molecular distribution represented in SMILES strings to improve molecule generation.

Table \ref{tab:pmo_tasks} presents the AUC-Top10 results across multiple PMO tasks. We observe that LSTM PSV-PPO consistently outperforms LSTM PPO across all reported tasks, demonstrating that incorporating Partial SMILES Validation significantly improves molecular generation. Notably, the improvement from LSTM PPO to LSTM PSV-PPO is particularly meaningful, as the performance of PPO-based models is strongly influenced by the quality of the pre-trained model. These findings emphasize the effectiveness of PSV-PPO in stabilizing training and enhancing exploration, enabling it to generate higher-quality molecules while maintaining strong validity constraints. For completeness, the full table of results is provided in the appendix.

\begin{table*}[ht]
\centering
\caption{Performance Comparison of Different Methods on PMO Tasks (AUC-Top10)}
\label{tab:pmo_tasks}
\begin{tabular}{@{}lcccc@{}}
\toprule
Task & REINVENT & LSTM HC & LSTM PPO & LSTM PSV-PPO \\ 
\midrule
albuterol\_similarity & \textbf{0.882±0.006} & 0.719±0.018  & 0.527±0.014 & 0.761±0.007 \\
amlodipine\_mpo & 0.635±0.035  & 0.593±0.016  & 0.587±0.008 & \textbf{0.647±0.007} \\
celecoxib\_rediscovery & \textbf{0.713±0.067}  & 0.539±0.018  & 0.532±0.041 & 0.612±0.021 \\
deco\_hop & 0.666±0.044  & \textbf{0.826±0.017}  & 0761±0.008 & 0.802±0.001 \\
drd2 & 0.945±0.007 & 0.919±0.015  & 0.883±0.012 & \textbf{0.959±0.013} \\
fexofenadine\_mpo & \textbf{0.784±0.006} & 0.725±0.003 & 0.695±0.003 & 0.698±0.001 \\
gsk3b & 0.865±0.043  & 0.839±0.015   & 0.794±0.023 & \textbf{0.869±0.089} \\
isomers\_c9h10n2o2pf2cl & 0.642±0.054 & 0.342±0.027  & 0.608±0.001 & \textbf{0.652±0.018} \\
\bottomrule
\end{tabular}
\end{table*}

\subsection{Docking}
Molecular docking provides an essential metric for evaluating the effectiveness of generative models in producing biologically relevant molecules. We evaluate our approach on docking tasks for a well-studied protein target: fa7\cite{yang2021hit}. The docking performance is assessed using multiple metrics, including AUC-1, AUC-10, AUC-100, synthetic accessibility (SA), and diversity (Div). 

To perform docking, we utilize the PyScreener framework with QVina2 as the docking backend, which enables efficient and accurate binding affinity predictions. As shown in Figure \ref{fig:docking}, PSV-PPO consistently achieves higher scores than PPO and HC across all evaluated metrics. This result demonstrates the effectiveness of PSV-PPO in optimizing molecular docking performance.

\subsection{GuacaMol Benchmark}
\subsubsection{Pretraining} We pre-trained our model on the ChEMBL dataset \cite{gaulton2012chembl}, available through the Guacamol benchmark \cite{brown2019guacamol}. For this, we fine-tuned a pre-trained GPT-2 model \cite{lagler2013gpt2} from Hugging Face \cite{wolf2020transformers} for five epochs. A linear warmup scheduler was utilized with a maximum learning rate of 0.0001. The atomwise tokenizer is provided by the SPE package \cite{li2021smiles}. To meet the requirement of atomwise tokenizer, the final fully connected layer of the model was resized to align with the 48-token vocabulary of our atom-wise tokenizer, which includes tokens for beginning-of-sequence (BOS), end-of-sequence (EOS), and padding (PAD) tokens. The GPT-2 model used contains 1.5 billion trainable parameters across 12 transformer layers, each with an embedding size of 768. 

We set the maximum SMILES sequences length to 100 tokens, though this can be adjusted per task to optimize GPU memory usage. Pretraining was conducted on a single NVIDIA A100 GPU for 8 hours. Upon completion, the pre-trained model achieved a validation rate of $0.9982$ and a uniqueness rate of $0.998$ across $30,000$ generated samples. This metrics align with typical outcomes for pretrained models \cite{polykovskiy2020molecular}.

\subsubsection{Benchmark Tasks}
The GuacaMol benchmark is a comprehensive, open-source evaluation framework designed to assess the performance of de novo molecular design algorithms. It features 20 goal-directed tasks that closely simulate real-world drug discovery challenges. These tasks are categorized into various objectives commonly pursued in drug design, such as drug rediscovery, structural similarity, physicochemical property optimization, etc. The tasks assess the models' ability to generate molecules with specific structural and physicochemical properties or to optimize multiple objectives simultaneously. The description of baseline models are explained in Appendix.
\begin{figure}[htbp]
\centering
\includegraphics[width=0.4\textwidth]{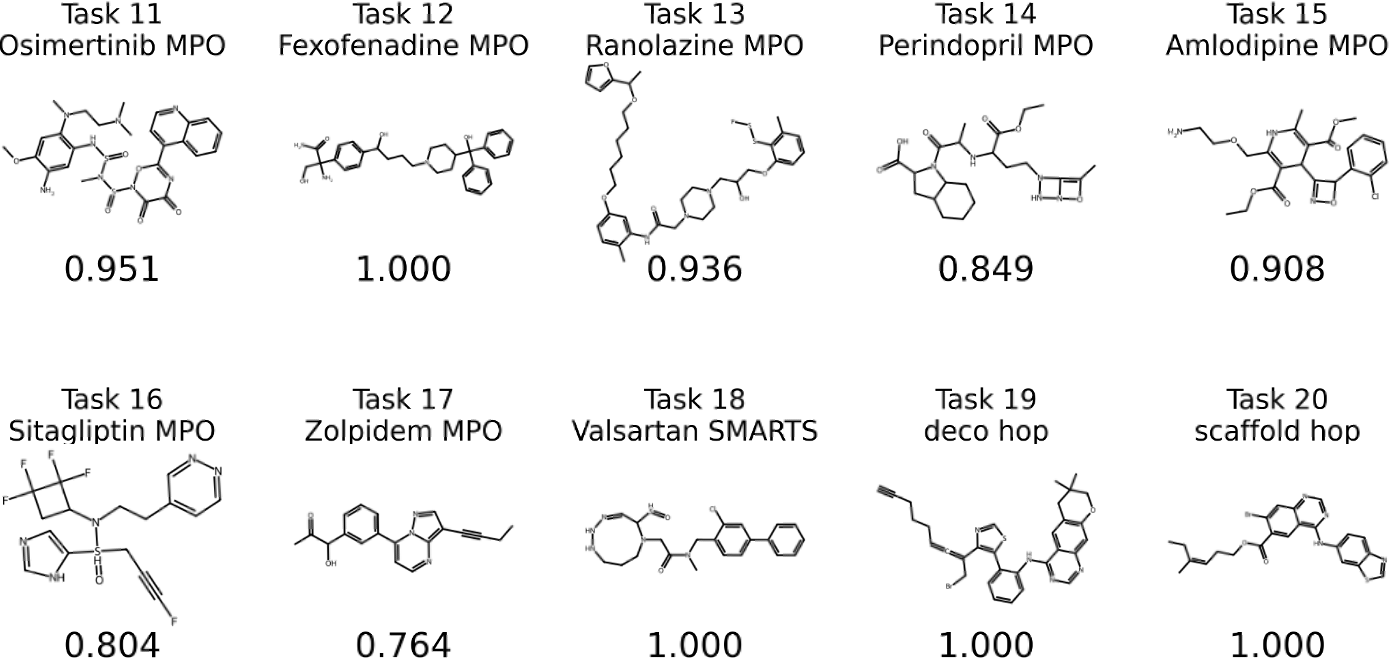}
\caption{Illustration of the top-scoring molecules generated by PSV-PPO for tasks 11-20 on the GuacaMol benchmark.} 

\label{fig:generated}
\end{figure}
The performance of our models was evaluated using the standard scoring functions provided by the GuacaMol benchmark. For each task, the models was trained for up to $500$ epochs, with an episode size of 512 and a batch size of 256. Notably, for the Valsartan SMARTS task, we observed that the initial molecule scores was zero, which could lead to aimless exploration by the model. To mitigate this, we implemented a warm-up sampling strategy. In this phase, the pretrained model was employed for generation until at least one molecule with a score greater than 0.01 was successfully discovered and added to the experience replay memory pool. Furthermore, we set a maximum tokenized SMILES sequence length of 100 tokens. Any sequences that did not produce an EOS token before reaching this length limit were discarded during generation.
\subsubsection{Benchmark Results}
The benchmark results in Table \ref{tab:benchmark_scores} follow the results reported in MoRL-MGPT \cite{hu2024novo}, allowing for direct comparison between our method and the baseline models, which are: (1) SMILES GA \cite{yoshikawa2018population}, (2) SMILES LSTM \cite{segler2018generating}, (3) Graph GA \cite{jensen2019graph}, (4) Reinvent \cite{olivecrona2017molecular}, (5) GEGL \cite{ahn2020guiding}, and (6) MoRL-MGPT. We observe that PSV-PPO outperforms existing baselines and achieves $14$ our of $20$ highest scores among $20$ tasks, including $11$ tasks that our method acheves $1.00$. This demonstrates the effectiveness of our methods. In Figure \ref{fig:generated}, we present the molecule with highest score from task 11 to task 20 with corresponding scores for each task. See Appendix for full illustrations of top 20 molecules and the diversity of generated molecules.

\subsection{Ablation Study}
To better understand the contribution of different components in PSV-PPO, we conducted ablation studies focusing on the use of PSV table, and the GPS and TPC losses. The results of these studies for the Fexofenadine MPO and Osimertinib MPO tasks are illustrated in Figure \ref{fig:ablation}.
\subsubsection{Contribution from the PSV Table}
The inclusion of the PSV table is critical for maintaining the validity rate of the generated molecules. As shown in the second column of Figure \ref{fig:ablation}, without the PSV table, the valid rate drastically drops below $10\%$ after a certain number of epochs, which indicates that the model frequently generates invalid SMILES strings. Moreover, we also observes that the policy loss starts to show high peaks and eventually end up with gradient explosion and caused all parameters become "nan". 

\subsubsection{Effect of Token Probability Control}
The GPS and TPC losses help prevent mode collapse by penalizing overconfident token selections. The Duplicated in Experience and Duplicated in Generation plots (third and fourth columns in Figure \ref{fig:ablation}) demonstrate that models without this control mechanism tend to generate more duplicates. This indicates a lack of diversity in the generated molecules, which directly impacts the effectiveness of the exploration process. In summary, the ablation studies confirm that each component of PSV-PPO contributes significantly to the overall performance. The PSV table ensures high validity, length normalization controls sequence length, and token probability control enhances diversity, all of which are crucial for effective and efficient molecular generation.

\section{Conclusion}
In this study, we introduced PSV-PPO, an innovative variant of the Proximal Policy Optimization (PPO) algorithm, specifically designed to address the challenges of reinforcement learning in SMILES-based deep generative models. By incorporating a partial SMILES validation scheme directly into the reinforcement learning process, PSV-PPO achieves outstanding performance on the GuacaMol benchmark, and shows improvement from PPO in all tasks on the PMO benchmark. Our ablation studies underscore the importance of each component within the framework, highlighting the critical role of the PSV-PPO in enhancing exploration while maintaining stability and validity.

The Partial SMILES Validation scheme is versatile and independent of the specific task at hand, making it compatible with any sequential model that utilizes the SMILES-based auto-regressive generation design. This adaptability suggests that PSV-PPO could have broad applications across various areas of de novo drug discovery, offering a significant advancement in the development of more reliable and effective generative models.

\bibliography{references}

\begin{thebibliography}{49}
\providecommand{\natexlab}[1]{#1}
\providecommand{\url}[1]{\texttt{#1}}
\expandafter\ifx\csname urlstyle\endcsname\relax
  \providecommand{\doi}[1]{doi: #1}\else
  \providecommand{\doi}{doi: \begingroup \urlstyle{rm}\Url}\fi

\bibitem[Ahn et~al.(2020)Ahn, Kim, Lee, and Shin]{ahn2020guiding}
Ahn, S., Kim, J., Lee, H., and Shin, J.
\newblock Guiding deep molecular optimization with genetic exploration.
\newblock \emph{Advances in neural information processing systems}, 33:\penalty0 12008--12021, 2020.

\bibitem[Andrychowicz et~al.(2020)Andrychowicz, Baker, Chociej, Jozefowicz, McGrew, Pachocki, Petron, Plappert, Powell, Ray, et~al.]{andrychowicz2020learning}
Andrychowicz, O.~M., Baker, B., Chociej, M., Jozefowicz, R., McGrew, B., Pachocki, J., Petron, A., Plappert, M., Powell, G., Ray, A., et~al.
\newblock Learning dexterous in-hand manipulation.
\newblock \emph{The International Journal of Robotics Research}, 39\penalty0 (1):\penalty0 3--20, 2020.

\bibitem[Arjovsky et~al.(2017)Arjovsky, Chintala, and Bottou]{arjovsky2017wasserstein}
Arjovsky, M., Chintala, S., and Bottou, L.
\newblock Wasserstein generative adversarial networks.
\newblock In \emph{International conference on machine learning}, pp.\  214--223. PMLR, 2017.

\bibitem[Ar{\'u}s-Pous et~al.(2020)Ar{\'u}s-Pous, Patronov, Bjerrum, Tyrchan, Reymond, Chen, and Engkvist]{arus2020smiles}
Ar{\'u}s-Pous, J., Patronov, A., Bjerrum, E.~J., Tyrchan, C., Reymond, J.-L., Chen, H., and Engkvist, O.
\newblock Smiles-based deep generative scaffold decorator for de-novo drug design.
\newblock \emph{Journal of cheminformatics}, 12:\penalty0 1--18, 2020.

\bibitem[Bagal et~al.(2021)Bagal, Aggarwal, Vinod, and Priyakumar]{bagal2021molgpt}
Bagal, V., Aggarwal, R., Vinod, P., and Priyakumar, U.~D.
\newblock Molgpt: molecular generation using a transformer-decoder model.
\newblock \emph{Journal of Chemical Information and Modeling}, 62\penalty0 (9):\penalty0 2064--2076, 2021.

\bibitem[Bjerrum \& Threlfall(2017)Bjerrum and Threlfall]{bjerrum2017molecular}
Bjerrum, E.~J. and Threlfall, R.
\newblock Molecular generation with recurrent neural networks (rnns).
\newblock \emph{arXiv preprint arXiv:1705.04612}, 2017.

\bibitem[Blaschke et~al.(2020)Blaschke, Ar{\'u}s-Pous, Chen, Margreitter, Tyrchan, Engkvist, Papadopoulos, and Patronov]{blaschke2020reinvent}
Blaschke, T., Ar{\'u}s-Pous, J., Chen, H., Margreitter, C., Tyrchan, C., Engkvist, O., Papadopoulos, K., and Patronov, A.
\newblock Reinvent 2.0: an ai tool for de novo drug design.
\newblock \emph{Journal of chemical information and modeling}, 60\penalty0 (12):\penalty0 5918--5922, 2020.

\bibitem[Brown et~al.(2019)Brown, Fiscato, Segler, and Vaucher]{brown2019guacamol}
Brown, N., Fiscato, M., Segler, M.~H., and Vaucher, A.~C.
\newblock Guacamol: benchmarking models for de novo molecular design.
\newblock \emph{Journal of chemical information and modeling}, 59\penalty0 (3):\penalty0 1096--1108, 2019.

\bibitem[Dhariwal et~al.(2017)Dhariwal, Hesse, Klimov, Nichol, Plappert, Radford, Schulman, Sidor, Wu, and Zhokhov]{baselines}
Dhariwal, P., Hesse, C., Klimov, O., Nichol, A., Plappert, M., Radford, A., Schulman, J., Sidor, S., Wu, Y., and Zhokhov, P.
\newblock Openai baselines.
\newblock \url{https://github.com/openai/baselines}, 2017.

\bibitem[Duvenaud et~al.(2015)Duvenaud, Maclaurin, Iparraguirre, Bombarell, Hirzel, Aspuru-Guzik, and Adams]{duvenaud2015convolutional}
Duvenaud, D.~K., Maclaurin, D., Iparraguirre, J., Bombarell, R., Hirzel, T., Aspuru-Guzik, A., and Adams, R.~P.
\newblock Convolutional networks on graphs for learning molecular fingerprints.
\newblock \emph{Advances in neural information processing systems}, 28, 2015.

\bibitem[Gao et~al.(2022)Gao, Fu, Sun, and Coley]{gao2022sample}
Gao, W., Fu, T., Sun, J., and Coley, C.
\newblock Sample efficiency matters: a benchmark for practical molecular optimization.
\newblock \emph{Advances in neural information processing systems}, 35:\penalty0 21342--21357, 2022.

\bibitem[Gaulton et~al.(2012)Gaulton, Bellis, Bento, Chambers, Davies, Hersey, Light, McGlinchey, Michalovich, Al-Lazikani, et~al.]{gaulton2012chembl}
Gaulton, A., Bellis, L.~J., Bento, A.~P., Chambers, J., Davies, M., Hersey, A., Light, Y., McGlinchey, S., Michalovich, D., Al-Lazikani, B., et~al.
\newblock Chembl: a large-scale bioactivity database for drug discovery.
\newblock \emph{Nucleic acids research}, 40\penalty0 (D1):\penalty0 D1100--D1107, 2012.

\bibitem[Ghugare et~al.(2023)Ghugare, Miret, Hugessen, Phielipp, and Berseth]{ghugare2023searching}
Ghugare, R., Miret, S., Hugessen, A., Phielipp, M., and Berseth, G.
\newblock Searching for high-value molecules using reinforcement learning and transformers.
\newblock \emph{arXiv preprint arXiv:2310.02902}, 2023.

\bibitem[Gilmer et~al.(2017)Gilmer, Schoenholz, Riley, Vinyals, and Dahl]{gilmer2017neural}
Gilmer, J., Schoenholz, S.~S., Riley, P.~F., Vinyals, O., and Dahl, G.~E.
\newblock Neural message passing for quantum chemistry.
\newblock In \emph{International conference on machine learning}, pp.\  1263--1272. PMLR, 2017.

\bibitem[G{\'o}mez-Bombarelli et~al.(2018)G{\'o}mez-Bombarelli, Wei, Duvenaud, Hern{\'a}ndez-Lobato, S{\'a}nchez-Lengeling, Sheberla, Aguilera-Iparraguirre, Hirzel, Adams, and Aspuru-Guzik]{gomez2018automatic}
G{\'o}mez-Bombarelli, R., Wei, J.~N., Duvenaud, D., Hern{\'a}ndez-Lobato, J.~M., S{\'a}nchez-Lengeling, B., Sheberla, D., Aguilera-Iparraguirre, J., Hirzel, T.~D., Adams, R.~P., and Aspuru-Guzik, A.
\newblock Automatic chemical design using a data-driven continuous representation of molecules.
\newblock \emph{ACS central science}, 4\penalty0 (2):\penalty0 268--276, 2018.

\bibitem[Hao et~al.(2023)Hao, Yang, Tang, Bai, Liu, Meng, Liu, and Wang]{hao2023exploration}
Hao, J., Yang, T., Tang, H., Bai, C., Liu, J., Meng, Z., Liu, P., and Wang, Z.
\newblock Exploration in deep reinforcement learning: From single-agent to multiagent domain.
\newblock \emph{IEEE Transactions on Neural Networks and Learning Systems}, 2023.

\bibitem[Hellinger(1909)]{hellinger1909neue}
Hellinger, E.
\newblock Neue begr{\"u}ndung der theorie quadratischer formen von unendlichvielen ver{\"a}nderlichen.
\newblock \emph{Journal f{\"u}r die reine und angewandte Mathematik}, 1909\penalty0 (136):\penalty0 210--271, 1909.

\bibitem[Henderson et~al.(2018)Henderson, Islam, Bachman, Pineau, Precup, and Meger]{henderson2018deep}
Henderson, P., Islam, R., Bachman, P., Pineau, J., Precup, D., and Meger, D.
\newblock Deep reinforcement learning that matters.
\newblock In \emph{Proceedings of the AAAI conference on artificial intelligence}, volume~32, 2018.

\bibitem[Hu et~al.(2024)Hu, Liu, Zhao, and Zhang]{hu2024novo}
Hu, X., Liu, G., Zhao, Y., and Zhang, H.
\newblock De novo drug design using reinforcement learning with multiple gpt agents.
\newblock \emph{Advances in Neural Information Processing Systems}, 36, 2024.

\bibitem[Irwin et~al.(2022)Irwin, Dimitriadis, He, and Bjerrum]{irwin2022chemformer}
Irwin, R., Dimitriadis, S., He, J., and Bjerrum, E.~J.
\newblock Chemformer: a pre-trained transformer for computational chemistry.
\newblock \emph{Machine Learning: Science and Technology}, 3\penalty0 (1):\penalty0 015022, 2022.

\bibitem[Jensen(2019)]{jensen2019graph}
Jensen, J.~H.
\newblock A graph-based genetic algorithm and generative model/monte carlo tree search for the exploration of chemical space.
\newblock \emph{Chemical science}, 10\penalty0 (12):\penalty0 3567--3572, 2019.

\bibitem[Jin et~al.(2018)Jin, Barzilay, and Jaakkola]{jin2018junction}
Jin, W., Barzilay, R., and Jaakkola, T.
\newblock Junction tree variational autoencoder for molecular graph generation.
\newblock In \emph{International conference on machine learning}, pp.\  2323--2332. PMLR, 2018.

\bibitem[Jin et~al.(2020)Jin, Barzilay, and Jaakkola]{jin2020hierarchical}
Jin, W., Barzilay, R., and Jaakkola, T.
\newblock Hierarchical generation of molecular graphs using structural motifs.
\newblock In \emph{International conference on machine learning}, pp.\  4839--4848. PMLR, 2020.

\bibitem[Kearnes et~al.(2016)Kearnes, McCloskey, Berndl, Pande, and Riley]{kearnes2016molecular}
Kearnes, S., McCloskey, K., Berndl, M., Pande, V., and Riley, P.
\newblock Molecular graph convolutions: moving beyond fingerprints.
\newblock \emph{Journal of computer-aided molecular design}, 30:\penalty0 595--608, 2016.

\bibitem[Kemker et~al.(2018)Kemker, McClure, Abitino, Hayes, and Kanan]{kemker2018measuring}
Kemker, R., McClure, M., Abitino, A., Hayes, T., and Kanan, C.
\newblock Measuring catastrophic forgetting in neural networks.
\newblock In \emph{Proceedings of the AAAI conference on artificial intelligence}, volume~32, 2018.

\bibitem[Kirkpatrick et~al.(2017)Kirkpatrick, Pascanu, Rabinowitz, Veness, Desjardins, Rusu, Milan, Quan, Ramalho, Grabska-Barwinska, et~al.]{kirkpatrick2017overcoming}
Kirkpatrick, J., Pascanu, R., Rabinowitz, N., Veness, J., Desjardins, G., Rusu, A.~A., Milan, K., Quan, J., Ramalho, T., Grabska-Barwinska, A., et~al.
\newblock Overcoming catastrophic forgetting in neural networks.
\newblock \emph{Proceedings of the national academy of sciences}, 114\penalty0 (13):\penalty0 3521--3526, 2017.

\bibitem[Korshunova et~al.(2022)Korshunova, Huang, Capuzzi, Radchenko, Savych, Moroz, Wells, Willson, Tropsha, and Isayev]{korshunova2022generative}
Korshunova, M., Huang, N., Capuzzi, S., Radchenko, D.~S., Savych, O., Moroz, Y.~S., Wells, C.~I., Willson, T.~M., Tropsha, A., and Isayev, O.
\newblock Generative and reinforcement learning approaches for the automated de novo design of bioactive compounds.
\newblock \emph{Communications Chemistry}, 5\penalty0 (1):\penalty0 129, 2022.

\bibitem[Kostrikov(2018)]{kostrikov2018pytorch}
Kostrikov, I.
\newblock Pytorch implementations of reinforcement learning algorithms, 2018.

\bibitem[Krenn et~al.(2022)Krenn, Ai, Barthel, Carson, Frei, Frey, Friederich, Gaudin, Gayle, Jablonka, et~al.]{krenn2022selfies}
Krenn, M., Ai, Q., Barthel, S., Carson, N., Frei, A., Frey, N.~C., Friederich, P., Gaudin, T., Gayle, A.~A., Jablonka, K.~M., et~al.
\newblock Selfies and the future of molecular string representations.
\newblock \emph{Patterns}, 3\penalty0 (10), 2022.

\bibitem[Kullback \& Leibler(1951)Kullback and Leibler]{kullback1951information}
Kullback, S. and Leibler, R.~A.
\newblock On information and sufficiency.
\newblock \emph{The annals of mathematical statistics}, 22\penalty0 (1):\penalty0 79--86, 1951.

\bibitem[Kusner et~al.(2017)Kusner, Paige, and Hern{\'a}ndez-Lobato]{kusner2017grammar}
Kusner, M.~J., Paige, B., and Hern{\'a}ndez-Lobato, J.~M.
\newblock Grammar variational autoencoder.
\newblock In \emph{International conference on machine learning}, pp.\  1945--1954. PMLR, 2017.

\bibitem[Lagler et~al.(2013)Lagler, Schindelegger, B{\"o}hm, Kr{\'a}sn{\'a}, and Nilsson]{lagler2013gpt2}
Lagler, K., Schindelegger, M., B{\"o}hm, J., Kr{\'a}sn{\'a}, H., and Nilsson, T.
\newblock Gpt2: Empirical slant delay model for radio space geodetic techniques.
\newblock \emph{Geophysical research letters}, 40\penalty0 (6):\penalty0 1069--1073, 2013.

\bibitem[Li \& Fourches(2021)Li and Fourches]{li2021smiles}
Li, X. and Fourches, D.
\newblock Smiles pair encoding: a data-driven substructure tokenization algorithm for deep learning.
\newblock \emph{Journal of chemical information and modeling}, 61\penalty0 (4):\penalty0 1560--1569, 2021.

\bibitem[Lo et~al.(2018)Lo, Rensi, Torng, and Altman]{lo2018machine}
Lo, Y.-C., Rensi, S.~E., Torng, W., and Altman, R.~B.
\newblock Machine learning in chemoinformatics and drug discovery.
\newblock \emph{Drug discovery today}, 23\penalty0 (8):\penalty0 1538--1546, 2018.

\bibitem[Mokaya et~al.(2023)Mokaya, Imrie, van Hoorn, Kalisz, Bradley, and Deane]{mokaya2023testing}
Mokaya, M., Imrie, F., van Hoorn, W.~P., Kalisz, A., Bradley, A.~R., and Deane, C.~M.
\newblock Testing the limits of smiles-based de novo molecular generation with curriculum and deep reinforcement learning.
\newblock \emph{Nature Machine Intelligence}, 5\penalty0 (4):\penalty0 386--394, 2023.

\bibitem[O'Boyle(2024)]{partialsmiles2024}
O'Boyle, N.
\newblock partialsmiles, version 1.0.
\newblock \url{https://github.com/baoilleach/partialsmiles}, 2024.

\bibitem[Olivecrona et~al.(2017)Olivecrona, Blaschke, Engkvist, and Chen]{olivecrona2017molecular}
Olivecrona, M., Blaschke, T., Engkvist, O., and Chen, H.
\newblock Molecular de-novo design through deep reinforcement learning.
\newblock \emph{Journal of cheminformatics}, 9:\penalty0 1--14, 2017.

\bibitem[Polykovskiy et~al.(2020)Polykovskiy, Zhebrak, Sanchez-Lengeling, Golovanov, Tatanov, Belyaev, Kurbanov, Artamonov, Aladinskiy, Veselov, et~al.]{polykovskiy2020molecular}
Polykovskiy, D., Zhebrak, A., Sanchez-Lengeling, B., Golovanov, S., Tatanov, O., Belyaev, S., Kurbanov, R., Artamonov, A., Aladinskiy, V., Veselov, M., et~al.
\newblock Molecular sets (moses): a benchmarking platform for molecular generation models.
\newblock \emph{Frontiers in pharmacology}, 11:\penalty0 565644, 2020.

\bibitem[Ramsundar et~al.(2019)Ramsundar, Eastman, Walters, Pande, Leswing, and Wu]{Ramsundar-et-al-2019}
Ramsundar, B., Eastman, P., Walters, P., Pande, V., Leswing, K., and Wu, Z.
\newblock \emph{Deep Learning for the Life Sciences}.
\newblock O'Reilly Media, 2019.
\newblock \url{https://www.amazon.com/Deep-Learning-Life-Sciences-Microscopy/dp/1492039837}.

\bibitem[Schulman et~al.(2015)Schulman, Levine, Abbeel, Jordan, and Moritz]{schulman2015trust}
Schulman, J., Levine, S., Abbeel, P., Jordan, M., and Moritz, P.
\newblock Trust region policy optimization.
\newblock In \emph{International conference on machine learning}, pp.\  1889--1897. PMLR, 2015.

\bibitem[Schulman et~al.(2017)Schulman, Wolski, Dhariwal, Radford, and Klimov]{schulman2017proximal}
Schulman, J., Wolski, F., Dhariwal, P., Radford, A., and Klimov, O.
\newblock Proximal policy optimization algorithms.
\newblock \emph{arXiv preprint arXiv:1707.06347}, 2017.

\bibitem[Segler et~al.(2018)Segler, Kogej, Tyrchan, and Waller]{segler2018generating}
Segler, M.~H., Kogej, T., Tyrchan, C., and Waller, M.~P.
\newblock Generating focused molecule libraries for drug discovery with recurrent neural networks.
\newblock \emph{ACS central science}, 4\penalty0 (1):\penalty0 120--131, 2018.

\bibitem[Shi et~al.(2018)Shi, Chen, Qiu, and Huang]{shi2018toward}
Shi, Z., Chen, X., Qiu, X., and Huang, X.
\newblock Toward diverse text generation with inverse reinforcement learning.
\newblock \emph{arXiv preprint arXiv:1804.11258}, 2018.

\bibitem[Weininger(1988)]{weininger1988smiles}
Weininger, D.
\newblock Smiles, a chemical language and information system. 1. introduction to methodology and encoding rules.
\newblock \emph{Journal of chemical information and computer sciences}, 28\penalty0 (1):\penalty0 31--36, 1988.

\bibitem[Wolf et~al.(2020)Wolf, Debut, Sanh, Chaumond, Delangue, Moi, Cistac, Rault, Louf, Funtowicz, et~al.]{wolf2020transformers}
Wolf, T., Debut, L., Sanh, V., Chaumond, J., Delangue, C., Moi, A., Cistac, P., Rault, T., Louf, R., Funtowicz, M., et~al.
\newblock Transformers: State-of-the-art natural language processing.
\newblock In \emph{Proceedings of the 2020 conference on empirical methods in natural language processing: system demonstrations}, pp.\  38--45, 2020.

\bibitem[Wu et~al.(2023)Wu, He, Liu, Sun, Liu, Han, and Tang]{wu2023brief}
Wu, T., He, S., Liu, J., Sun, S., Liu, K., Han, Q.-L., and Tang, Y.
\newblock A brief overview of chatgpt: The history, status quo and potential future development.
\newblock \emph{IEEE/CAA Journal of Automatica Sinica}, 10\penalty0 (5):\penalty0 1122--1136, 2023.

\bibitem[Wu et~al.(2018)Wu, Ramsundar, Feinberg, Gomes, Geniesse, Pappu, Leswing, and Pande]{wu2018moleculenet}
Wu, Z., Ramsundar, B., Feinberg, E.~N., Gomes, J., Geniesse, C., Pappu, A.~S., Leswing, K., and Pande, V.
\newblock Moleculenet: a benchmark for molecular machine learning.
\newblock \emph{Chemical science}, 9\penalty0 (2):\penalty0 513--530, 2018.

\bibitem[Yang et~al.(2021)Yang, Hwang, Lee, Ryu, and Hwang]{yang2021hit}
Yang, S., Hwang, D., Lee, S., Ryu, S., and Hwang, S.~J.
\newblock Hit and lead discovery with explorative rl and fragment-based molecule generation.
\newblock \emph{Advances in Neural Information Processing Systems}, 34:\penalty0 7924--7936, 2021.

\bibitem[Yoshikawa et~al.(2018)Yoshikawa, Terayama, Sumita, Homma, Oono, and Tsuda]{yoshikawa2018population}
Yoshikawa, N., Terayama, K., Sumita, M., Homma, T., Oono, K., and Tsuda, K.
\newblock Population-based de novo molecule generation, using grammatical evolution.
\newblock \emph{Chemistry Letters}, 47\penalty0 (11):\penalty0 1431--1434, 2018.

\end{thebibliography}
\bibliographystyle{icml2024}
\newpage
\appendix
\onecolumn
\section{Technical and Implementation Details}
\begin{algorithm}[H]
\caption{Partial Validity Table Generator from SMILES Actions}
\label{alg:pmvgrid}
\textbf{Input}: Action Sample $A$, Vocabulary Decoder $D$, Index of First EOS $idx_{eos}$, EOS Token $tok_{eos}$, Token ID $id_{tok}$\\
\textbf{Output}: PSV Map $PSV$
\begin{algorithmic}[1] 
\STATE Initialize $PSV$ as a zero matrix of size $|A| \times |D|$.
\STATE Set $smi_{current} = \emptyset$.
\STATE Set $index_{invalid} = -1$.
\FOR{each $i, token_{action}$ in enumerate $A[:idx_{eos}+1]$}
    \FOR{each $j, token_{vocab}$ in enumerate $D$}
        \STATE $mol = None$
        \IF{$token_{vocab} == tok_{eos}$}
            \IF{$i > 10$}
                \STATE Continue
            \ENDIF
            \STATE $mol = CheckValid_{Completed}(smi_{current})$
        \ELSE
            \STATE $mol = CheckValid_{OnTheFly}(smi_{current} + token_{vocab})$
        \ENDIF
        \IF{$mol \neq None$}
            \STATE Set $PSV[i, j] = 1$
        \ELSE
            \STATE Set $PSV[i, j] = 0$
            \IF{$token_{action} == token_{vocab}$}
                \STATE Set $index_{invalid} = i$
            \ENDIF
        \ENDIF
    \ENDFOR
    \IF{$index_{invalid} \neq -1$}
        \IF{$\sum PSV[i, :] == 0$}
            \STATE Set $PSV[i, id_{tok}] = 1$
            \STATE \textbf{return} $PSV, index_{invalid}$
        \ENDIF
    \ENDIF
    \IF{$token_{action} == tok_{eos}$}
        \STATE \textbf{return} $PSV, index_{invalid}$
    \ENDIF
    \STATE Append $token_{action}$ to $smi_{current}$
\ENDFOR
\STATE \textbf{return} $PSV, index_{invalid}$
\end{algorithmic}
\end{algorithm}
\subsection{PSV Table Generator Algorithm}
The algorithm presented in Algorithm \ref{alg:pmvgrid} outlines the process for generating a PSV table. The algorithm takes as input an action sample $A$, a vocabulary decoder $D$, the index of the first EOS token $idx_{eos}$, the EOS token itself $tok_{eos}$, and a token ID $id_{tok}$. The output is the PSV table, denoted as $PSV$, which is a matrix that captures the validity of each potential action at every step of the SMILES generation process.

The algorithm starts by initializing the PSV table as a zero matrix. A string variable $smi_{current}$ holds the constructed SMILES string, and $index_{invalid}$ tracks the first invalid index, initialized to -1. For each token in $A$ up to the EOS index, the algorithm checks the validity of appending each token from the vocabulary to $smi_{current}$. If the token is valid, the PSV entry is set to 1; otherwise, it is set to 0, and $index_{invalid}$ is updated if the invalid token matches the action token. If an entire row in the PSV table is invalid, the entry corresponding to $id_{tok}$ is forced to 1, and the algorithm returns the PSV table and the invalid index.

If the resulting molecular structure is valid, the corresponding entry in the PSV table is set to 1. If it is invalid, the entry is set to 0, and the index of the invalid action is recorded if the current action token matches the invalid vocabulary token. The algorithm then checks if all entries in the current row of the PSV table are zero. If so, it forces a valid action by setting the entry corresponding to $id_{tok}$ to 1 and returns the PSV table along with the invalid index. The loop continues until the EOS token is reached, at which point the PSV table and invalid index are returned.

This algorithm ensures that at each step of the SMILES generation process, the validity of the generated string is maintained by providing immediate feedback on the feasibility of each token addition, thereby guiding the model toward generating valid molecular structures.

\subsection{Adjusted Reward for Invalid SMILES Samples}
In reinforcement learning, delayed rewards can hinder the model's ability to associate specific actions with their outcomes. In the context of SMILES generation, assigning a negative reward only at the EOS token can obscure the source of invalidity, making it difficult for the model to learn which specific action caused the issue. To address this, we adjust the reward mechanism by penalizing the first invalid action as soon as it occurs, rather than waiting until the EOS token.

As detailed in Algorithm \ref{alg:pmvgrid}, if $index_{invalid} \neq -1$, a score of $-1$ is assigned immediately at this position, providing clear feedback to the model about the exact point of failure. This approach contrasts with traditional methods that apply penalties at the EOS token, thus allowing the model to quickly learn and avoid invalid token sequences.

The implementation of this adjusted reward system involves:

\begin{enumerate}
    \item \textbf{Identifying the Invalid Index:} After generating the PSV Table, if $index_{invalid}$ is not -1, it indicates where the SMILES string became invalid.
    \item \textbf{Assigning Immediate Penalty:} A score of -1 is placed at $index_{invalid}$, rather than at the EOS token, ensuring that the model is penalized at the exact moment of invalidity.
    \item \textbf{Standard Reward Handling:} If no invalid index is found ($index_{invalid} = -1$), reward assignment follows the usual protocol based on the task-specific reward function.
\end{enumerate}
This method provides immediate, clear feedback, helping the model better understand and correct errors in SMILES generation. By correcting invalid steps as soon as they occur, the model improves in generating valid molecular structures, making the learning process more effective overall.

\section{Loss Term Explanation}
\subsubsection{Global Probability Stabilization Loss (GPS Loss)}
During training, we observe that the model tends to assign increasingly high probabilities to specific molecular structures that yield high rewards. This behavior can lead to the generation of repetitive molecules, reducing diversity and diminishing the utility of the generated samples across epochs. To mitigate this issue, we introduce the Global Probability Stabilization Loss (GPS Loss). This loss penalizes the model when the cumulative probability of generating a particular sequence exceeds a predefined threshold, preventing the model from becoming overly biased toward a narrow subset of molecules.

The loss is computed as follows:

\begin{equation}
    L^{\text{GPS}}_{PSV}(\theta) = \frac{1}{\sqrt{2}} \left( \sqrt{\max\left(\prod_{t=1}^{T} \pi_{\theta}(a_t | s_t), \tau_{gps}\right)} - \sqrt{\tau_{gps}} \right)^2
\end{equation}

where:
\begin{itemize}
    \item \( \prod_{t=1}^{T} \pi_{\theta}(a_t | s_t) \) represents the cumulative probability of the sequence generated by the model.
    \item \( \tau_{gps} \) is the cumulative probability threshold (e.g., \( 1 \times 10^{-5} \)).
    \item \( \max(\cdot, \tau_{gps}) \) ensures that the cumulative probability is not penalized below the threshold.
\end{itemize}

This loss term encourages the model to maintain balanced exploration of the chemical space by discouraging excessive focus on any single molecular structure. By doing so, we aim to enhance the diversity of generated molecules and improve the overall quality of samples produced during training.

\subsubsection{Token Probability Control Loss (TPC Loss)}
To ensure the diversity of generated molecules and prevent the model from overcommitting to specific tokens, we introduce the Token Probability Control Loss (TPC Loss). This loss penalizes action token probabilities that exceed a predefined threshold \( \tau_{token} \), thereby discouraging the model from assigning disproportionately high probabilities to certain molecular fragments. The loss is computed as follows:

\begin{equation}
\begin{split}
L^{\text{TPC}}_{PSV}(\theta) &= \frac{1}{\sqrt{2}} \sum_{t=1}^{T} \left( \sum_{a \in A_t} \left[ \sqrt{\pi_{\theta}(a|s_t)} - \sqrt{\tau_{token}} \right]^2 \right.\\
&\quad \left. \cdot \frac{1}{|P_{PSV}(s_t)|} \right) \cdot \mathbb{I}(\pi_{\theta}(a|s_t) > \tau_{token})
\end{split}
\end{equation}

where:
\begin{itemize}
    \item \( \pi_{\theta}(a|s_t) \) is the probability assigned to action \( a \) given state \( s_t \).
    \item \( \tau_{token} \) is the predefined threshold for token probabilities.
    \item \( |P_{PSV}(s_t)| \) represents the number of valid candidate tokens at state \( s_t \).
    \item \( \mathbb{I}(\cdot) \) is an indicator function that activates the penalty only when \( \pi_{\theta}(a|s_t) > \tau_{token} \).
\end{itemize}

This formulation accounts for the size of the valid token set directly, ensuring that the penalty is appropriately scaled according to the number of choices available at each state.

\subsubsection{Length Normalization}
During our experiments, we observed that when the model stabilizes at a certain score level for an extended period, it tends to generate progressively longer sequences in an effort to explore uncharted chemical space. While this exploration is generally beneficial, excessively long sequences are often less relevant and can detract from the quality of the generated molecules. To mitigate this issue, we implement a length normalization term on the entropy loss and the token probability loss:

\begin{equation}
    L^{\text{LN}}_T = L_T \times \exp(-\gamma_{LN} \times T)
\end{equation}

This adjustment reduces the influence of excessively long sequences, guiding the model towards generating more plausible and relevant molecular structures.

\begin{table*}[h]
\centering
\begin{tabular}{|c|c|l|}
\hline
\textbf{Task ID} & \textbf{score} & \textbf{SMILES} \\
\hline
9  & 0.406 & CC1CCC(C(C)C2CCC(CC2=O)C3(C)C(C)C(O)C1 \\
10 & 0.400 & CCOc1cc(S(=O)(=O)N2CCN(CC2)cc1C1Cc2c([nH]c3ccccc23)C2CCC3c(C2)OCO3)N1C \\
11 & 0.951 & COc1cc(NC(C)(NS(=O)(=O)N(S(=O)=O)n2oc(-c3ccnc4cccc34)nc(=O)c2=O)cc1N1C \\
12 & 1.000 & O=C(O)(C(O)(C)C(C)CCN2CCC(C(O)(c3ccccc3)C3CC(C(O)C2)cc1 \\
13 & 0.936 & Cc1ccc(OCCCCCCOC(C)C2cco2)C1NC(=O)CN1CCN(C(C)(C)C2=O)C2CCS \\
14 & 0.849 & CCOC(=O)C(CCn1[nH]n2oc(C)c12)NC(C)C(=O)N1C(C(=O)O)CC2CCCCC21 \\
15 & 0.908 & CCOC(=O)C1=C(COCCN)NC(C)=C(C(=O)OC)C1C1=NOC1c1ccccc1Cl \\
16 & 0.804 & FC\#CCS(=O)(c1cnc[nH]1)N(CCc1ccnnc1)C1CC(F)(F)C1(F)F \\
17 & 0.764 & CCC\#Cc1cnn2c(-c3cccc(C(O)C(C)=O)c3)ccnc12 \\
18 & 0.999 & C([H])N(Cc1ccc(-c2ccccc2)cc1Cl)C(=O)CN1CCCNNN=CC1S(=O) \\
19 & 1.000 & C\#CCCCC=C=C(CBr)c1ncsc1-c1ccccc1Nc1ncnc2cc3c(cc12)CC(C)(C)CO3 \\
20 & 1.000 & CCC(C)=CCCOC(=O)c1cc2c(Nc3ccc4ncsc4c3)ncnc2cc1Br \\
\hline
\end{tabular}
\caption{SMILES Data with Scores}
\label{tab:smiles_data}
\end{table*}
\section{GuacaMol Results}
\subsection{Top-Scoring Molecules} 
We present visualizations of the top-scoring molecules generated by our model for osimertinib MPO and Amlodipine MPO. Notably, even when the top-20 molecules have very close scores, our results demonstrate that the generated molecules exhibit distinct structural differences. This contrasts with models that tend to produce molecules with similar structures when scores are close. Our model’s ability to generate structurally diverse molecules, even with minimal score variation, highlights its effectiveness in exploring different chemical spaces while maintaining high-quality outputs.
\begin{figure}[htbp]
\centering
\includegraphics[width=0.45\textwidth]{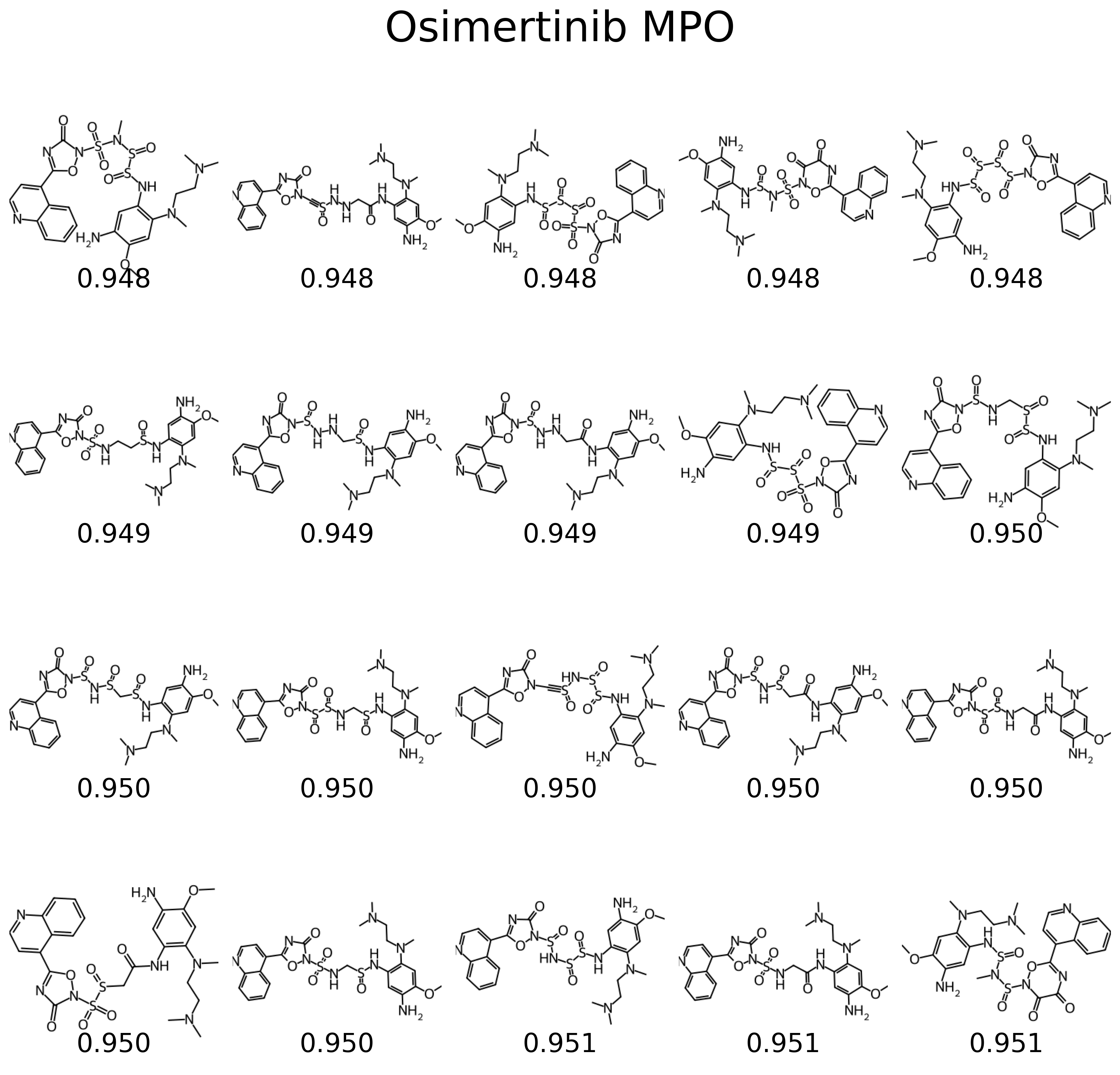}
\label{fig:generated}
\end{figure}
\begin{figure}[htbp]
\centering
\includegraphics[width=0.48\textwidth]{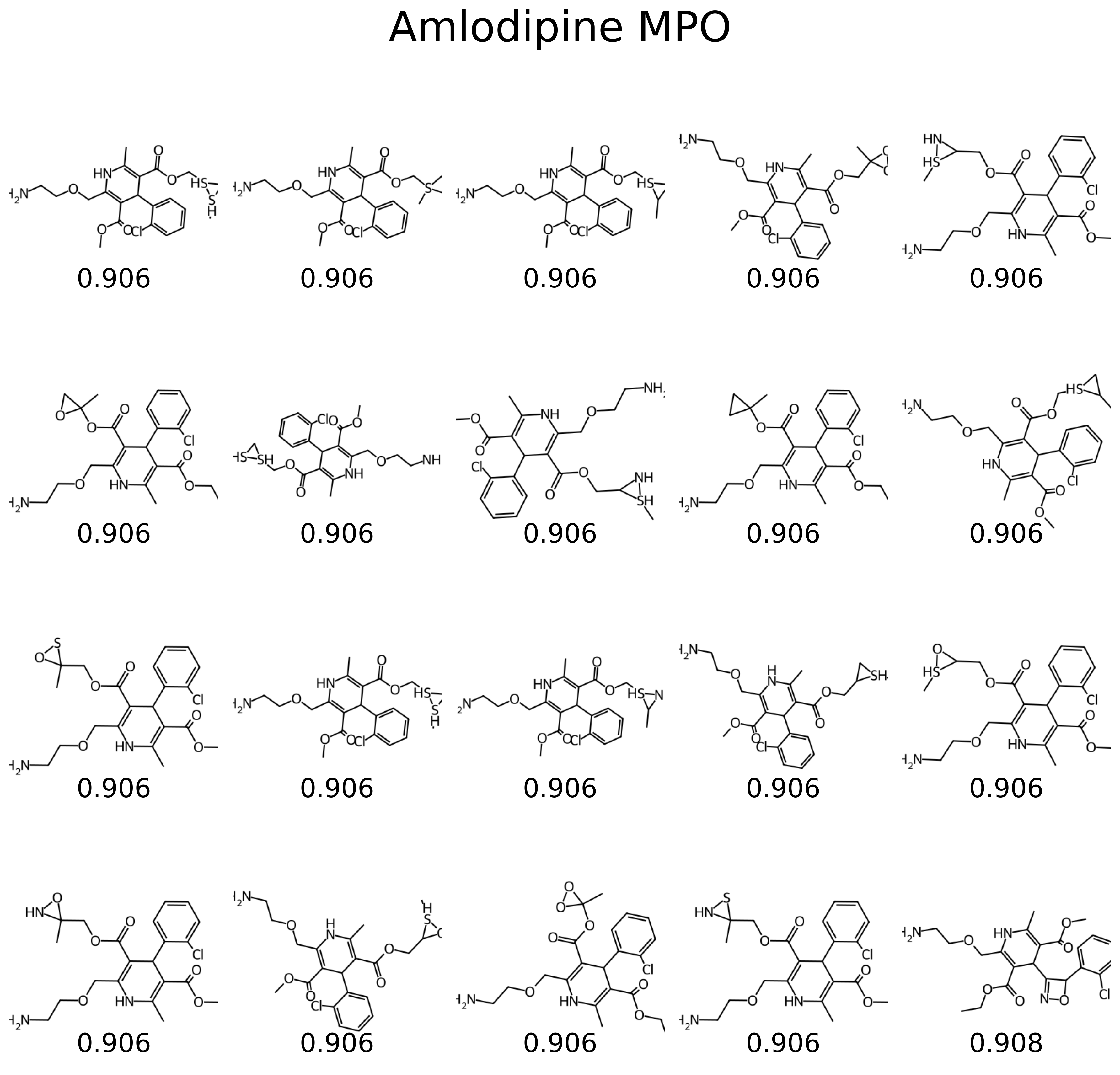}
\label{fig:generated}
\end{figure}

\begin{table*}[h]
\centering
\caption{Scores of PSV-PPO and other baseline models on the GuacaMol benchmark.}
\label{tab:benchmark_scores}
\footnotesize  
\begin{tabularx}{\textwidth}{lcccccccc}
\hline
\hline
\textbf{Tasks}                        & \textbf{SMILES GA} & \textbf{SMILES LSTM} & \textbf{Graph GA} & \textbf{Reinvent} & \textbf{GEGL} & \textbf{MolRL-MGPT} & \textbf{PSV-PPO} \\ \hline
1. Celecoxib rediscovery              & 0.732              &\textbf{1.000}               &\textbf{1.000}            &\textbf{1.000}            &\textbf{1.000}        & \textbf{1.000}               & \textbf{1.000}                \\ 
2. Troglitazone rediscovery           & 0.515              &\textbf{1.000}               &\textbf{1.000}            &\textbf{1.000}            & 0.552         & \textbf{1.000}               & \textbf{1.000}                \\ 
3. Thiothixene rediscovery            & 0.598              &\textbf{1.000}               &\textbf{1.000}            &\textbf{1.000}            &\textbf{1.000}        & \textbf{1.000}               & \textbf{1.000}                \\ 
4. Aripiprazole similarity            & 0.834              &\textbf{1.000}               &\textbf{1.000}            &\textbf{1.000}            &\textbf{1.000}        &\textbf{1.000}              & \textbf{1.000}                \\ 
5. Albuterol similarity               & 0.907              &\textbf{1.000}               &\textbf{1.000}            &\textbf{1.000}            &\textbf{1.000}        &\textbf{1.000}              & \textbf{1.000}                \\ 
6. Mestranol similarity               & 0.790              & \textbf{1.000}               & \textbf{1.000}             &\textbf{1.000}            &\textbf{1.000}        &\textbf{1.000}              & \textbf{1.000}                \\ 
7. C\textsubscript{11}H\textsubscript{24}                 & 0.829              & 0.993                & 0.971             & 0.999             &\textbf{1.000}        & \textbf{1.000}               & \textbf{1.000}                \\ 
8. C\textsubscript{9}H\textsubscript{10}N\textsubscript{2}O\textsubscript{2}P\textsubscript{2}Cl  & 0.889              & 0.879                & 0.982             & 0.877             &\textbf{1.000}        & 0.939               & \textbf{1.000}                \\ \hline
9. Median molecules 1                & 0.334              & 0.438                & 0.406             & 0.434             & 0.455         & 0.449               & \textbf{0.459}                \\ 
10. Median molecules 2               & 0.380              & 0.422                & 0.432             & 0.395             & \textbf{0.437}         & 0.422               & 0.392                \\ \hline
11. Osimertinib MPO                  & 0.886              & 0.907                & 0.953             & 0.889             &\textbf{1.000}        & 0.977               & 0.951                \\ 
12. Fexofenadine MPO                 & 0.931              & 0.959                & 0.998             &\textbf{1.000}            &\textbf{1.000}        &\textbf{1.000}              & \textbf{1.000}                \\ 
13. Ranolazine MPO                   & 0.881              & 0.855                & 0.920             & 0.895             & 0.933         & \textbf{0.939}               & 0.936               \\ 
14. Perindopril MPO                  & 0.661              & 0.808                & 0.792             & 0.764             & 0.833         & 0.810               & \textbf{0.849}                  \\ 
15. Amlodipine MPO                   & 0.722              & 0.894                & 0.894             & 0.888             & 0.905         & 0.906               & \textbf{0.908}                \\ 
16. Sitagliptin MPO                  & 0.689              & 0.545                & \textbf{0.891}             & 0.539             & 0.749         & 0.823               & 0.804                \\ 
17. Zaleplon MPO                     & 0.413              & 0.669                & 0.754             & 0.590             & 0.763         & \textbf{0.790}               & 0.764                \\ \hline
18. Valsartan SMARTS                 & 0.552              & 0.978                & 0.990             & 0.095             &\textbf{1.000}        & 0.997               & 0.999                \\ \hline
19. Deco hop                         & 0.970              & 0.996                &\textbf{1.000}            & 0.994             &\textbf{1.000}        &\textbf{1.000}              & \textbf{1.000}                \\ 
20. Scaffold hop                     & 0.885              & 0.998                &\textbf{1.000}            & 0.990             &\textbf{1.000}        &\textbf{1.000}              & \textbf{1.000}                \\ \hline
\textbf{Total}                       & 14.396             & 17.340               & 17.983            & 16.350            & 17.627        & 18.052              & \textbf{18.061}       \\
\hline
\hline
\end{tabularx}
\end{table*}

\begin{table*}[ht]
\centering
\caption{Performance Comparison of Different Methods on PMO Tasks (AUC-Top10)}
\label{tab:pmo_tasks_full}
\begin{tabular}{@{}lcccc@{}}
\toprule
Task & REINVENT & LSTM HC & LSTM PPO & LSTM PSV-PPO \\ 
\midrule
albuterol\_similarity & \textbf{0.882±0.006} & 0.719±0.018  & 0.527±0.014 & 0.761±0.007 \\
amlodipine\_mpo & 0.635±0.035  & 0.593±0.016  & 0.587±0.008 & \textbf{0.647±0.007} \\
celecoxib\_rediscovery & \textbf{0.713±0.067}  & 0.539±0.018  & 0.532±0.041 & 0.612±0.021 \\
deco\_hop & 0.666±0.044  & \textbf{0.826±0.017}  & 0761±0.008 & 0.802±0.001 \\
drd2 & 0.945±0.007 & 0.919±0.015  & 0.883±0.012 & \textbf{0.959±0.013} \\
fexofenadine\_mpo & \textbf{0.784±0.006} & 0.725±0.003 & 0.695±0.003 & 0.698±0.001 \\
gsk3b & 0.865±0.043  & 0.839±0.015   & 0.794±0.023 & \textbf{0.869±0.089} \\
isomers\_c7h8n2o2 & \textbf{0.852±0.036}  & 0.485±0.045  & 0.582±0.134 & 0.650±0.031 \\
isomers\_c9h10n2o2pf2cl & 0.642±0.054 & 0.342±0.027  & 0.608±0.001 & \textbf{0.652±0.018} \\
jnk3 & \textbf{0.783±0.023}  & 0.661±0.039  & 0.567±0.020 & 0.681±0.011 \\
median1 & \textbf{0.356±0.009}  & 0.255±0.010  & 0.204±0.004 & 0.193±0.003 \\
median2 & \textbf{0.276±0.008}  & 0.248±0.008 & 0.226±0.010 & 0.239±0.037 \\
mestranol\_similarity & \textbf{0.618±0.048} & 0.526±0.032  & 0.440±0.012 & 0.560±0.013 \\
osimertinib\_mpo & 0.837±0.009  & 0.796±0.002 & 0.777±0.153 & \textbf{0.844±0.008} \\
perindopril\_mpo & \textbf{0.537±0.016} & 0.489±0.007  & 0.461±0.019 & 0.452±0.002 \\
qed & 0.941±0.000  & 0.939±0.000 & 0.938±0.003 & 0.938±0.051 \\
\bottomrule
\end{tabular}
\end{table*}
\section{Full Benchmark Tables}

\end{document}